\documentclass[runningheads]{llncs}
\usepackage[T1]{fontenc}
\usepackage{graphicx}
\usepackage{amsmath}
\usepackage{amssymb}
\usepackage{mathtools}
\usepackage{microtype}
\usepackage{subfigure}
\usepackage{booktabs} 
\usepackage{xcolor}
\usepackage{bbold}
\DeclareMathOperator{\Tr}{Tr}
\begin{document}
\title{MaxCorrMGNN: A Multi-Graph Neural Network Framework for Generalized Multimodal Fusion of Medical Data for Outcome Prediction}

%
\author{Niharika S. D'Souza\inst{1}  \and Hongzhi Wang \inst{1} \and Andrea Giovannini \inst{2} \and Antonio Foncubierta-Rodriguez \inst{2} \and Kristen L. Beck \inst{1} \and Orest Boyko \inst{3} \and Tanveer Syeda-Mahmood \inst{1}}
%

\authorrunning{N.S. D'Souza et al.}
%
\institute{IBM Research Almaden, San Jose, CA, USA \and IBM Research, Zurich, Switzerland \and Department of Radiology,
VA Southern Nevada Healthcare System, NV, USA}

\maketitle              
\begin{abstract}
With the emergence of multimodal electronic health records, the evidence for an outcome may be captured across multiple modalities ranging from clinical to imaging and genomic data. Predicting outcomes effectively requires fusion frameworks capable of modeling fine-grained and multi-faceted complex interactions between modality features within and across patients. We develop an innovative fusion approach called MaxCorr MGNN that models non-linear modality correlations within and across patients through Hirschfeld-Gebelein-Re\`nyi maximal correlation (MaxCorr) embeddings, resulting in a multi-layered graph that preserves the identities of the modalities and patients.  We then design, for the first time, a generalized multi-layered graph neural network (MGNN) for task-informed reasoning in multi-layered graphs, that learns the parameters defining patient-modality graph connectivity and message passing in an end-to-end fashion. We evaluate our model an outcome prediction task on a Tuberculosis (TB) dataset consistently outperforming several state-of-the-art neural, graph-based and traditional fusion techniques.
\keywords{Multimodal Fusion \and Hirschfeld-Gebelein-Re\`nyi (HGR) maximal correlation \and Multi-Layered Graphs \and Multi-Graph Neural Networks}
\end{abstract}

\section{Introduction}
In the age of modern medicine, it is now possible to capture information about a patient through multiple data-rich modalities to give a holistic view of a patient's condition.  In complex diseases such as cancer~\cite{subramanian2020multimodal}, tuberculosis~\cite{munoz2010factors} or autism spectrum disorder~\cite{d2020deep,d2021deep,d2021matrix}, evidence for a diagnosis or treatment outcome may be present in multiple modalities such as clinical, genomic, molecular, pathological and radiological imaging. Reliable patient-tailored outcome prediction requires fusing information from modality data both within and across patients. This can be achieved by effectively modeling the fine-grained and multi-faceted complex interactions between modality features. In general, this is a challenging problem as it is largely unclear what information is best captured by each modality, how best to combine modalities, and how to effectively extract predictive patterns from data~\cite{lahat2015multimodal}.

\subsection{Related Works}
Existing attempts to fuse modalities for outcome prediction can be divided into at least three approaches, namely, feature vector-based, statistical or graph-based approaches. The vector-based approaches perform early, intermediate, or late fusion ~\cite{subramanian2020multimodal,baltruvsaitis2018multimodal} with the late fusion approach combining the results of prediction rather than fusing the modality features. Due to the restrictive nature of the underlying assumptions, these are often inadequate for characterizing the broader range of relationships among modality features and their relevance for prediction. In statistical approaches, methods such as canonical correlation analysis~\cite{subramanian2021multi} and its deep learning variants~\cite{yang2019survey} directly model feature correlations either in the native representation or in a latent space~\cite{subramanian2020multimodal}. However, these are not guaranteed to learn discriminative patterns in the unsupervised setting and can suffer from scalability issues when integrated into larger predictive models~\cite{wang2019efficient}. Recently, graph-based approaches have been developed which form basic~\cite{cosmo2020latent,zheng2022multi,dsouza2021m} or multiplexed graphs~\cite{d2022fusing} from latent embeddings derived from modality features using concatenation~\cite{cosmo2020latent} or weighted averaging~\cite{zheng2022multi}. Task-specific fusion is then achieved through inference via message passing walks between nodes in a graph neural network. In the basic collapsed graph construction, the inter-patient and intra-patient modality correlations are not fully distinguished. Conversely, in the multiplexed formulation~\cite{d2022fusing}, only a restricted form of multi-relational dependence is captured between nodes through vertical connections. Since the graph is defined using latent embedding directions, the modality semantics are not preserved. Additionally, the staged training of the graph construction and inference networks do not guarantee that the constructed graphs retain discriminable interaction patterns. 
\subsection{Our Contributions}
We develop a novel end-to-end fusion framework that addresses the limitations mentioned above. The \underline{M}aximal \underline{Corr}elation \underline{M}ulti-Layered \underline{G}raph \underline{N}eural \underline{N}etwork, i.e. MaxCorrMGNN, is a general yet interpretable framework for problems of multimodal fusion with unstructured data. Specifically, our approach marries the design principles of statistical representation learning with deep learning models for reasoning from multi-graphs. 

\begin{figure}[t!]
\begin{center}
\centerline{{\includegraphics[width=\textwidth]{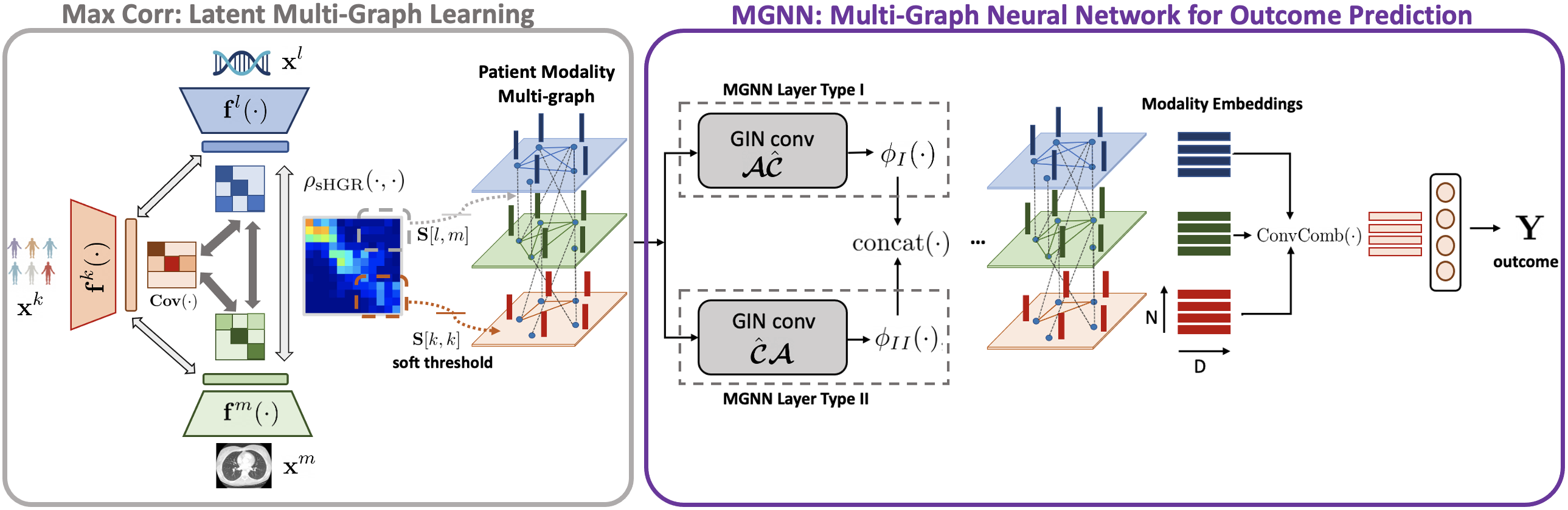}}}
\caption{{Our Generalized Framework for Multimodal Fusion \textbf{Gray Box:} The features from different modalities are input to the MaxCorr (HGR) formulation. The nodes are the patients of the Multi-Graph, and the planes are the modalities. \textbf{Purple Box:} The Multi-Graph Neural Network maps the multi-graph representation to the targets.}}
\label{MaxCorrMGNN_Formulation}
\end{center}
\end{figure}

\medskip
\noindent The main contributions of this work are three-fold:
\begin{itemize}
\item First, we propose to model intra and inter-patient modality relationships explicitly through a novel patient-modality multi-layered graph as shown in Fig.~\ref{MaxCorrMGNN_Formulation}. The edges in each layer (plane of the multi-graph) capture the \textit{intra-modality relations} between patients, while the cross-edges between layers capture \textit{inter-modality relations} across patients. 

\item Since these relationships are not known apriori for unstructured data, we propose, for the first time, to use learnable Hirschfeld-Gebelein-Re\`nyi (HGR) maximal correlations. We introduce learnable soft-thresholding to uncover salient connectivity patterns automatically. Effectively, this procedure allows us to express any multimodal dataset as a patient-modality multilayered graph for fusion. 

\item Third, we develop a multilayered graph neural network (MGNN) from first principles for task-informed reasoning from multi-layered graphs. 
\end{itemize}

\noindent To demonstrate the generality of our approach, we evaluate our framework on a large Tuberculosis (TB) dataset for multi-outcome prediction. Through rigorous experimentation, we show our framework outperforms several state-of-the-art graph based and traditional fusion baselines. 
\section{MaxCorrMGNN formulation for Multimodal Fusion}
We now describe the four main aspects of our formulation, namely, (a) multilayered graph representation, (b) formalism for maximal correlation (c) task-specific inference through graph neural networks, and (d) loss function for end-to-end learning of both graph connectivity and inference. 

\subsection{Patient-Modality Multi-layered Graph:} Given multimodal data about patients, we model the modality and patient information through a multi-layered graph~\cite{Cozzo2018} as shown in Fig.~\ref{MaxCorrMGNN_Formulation}. \textit{Here the nodes are grouped into multiple planes, each plane representing edge-connectivity according to an individual modality while each patient is represented by a set of corresponding nodes across the layers} (called a supra-node).

Mathematically, we represent the multi-layered graph as: $\mathcal{G}_{\text{M}} = (\mathcal{V},\mathcal{E}_{\text{M}})$, where $\vert{\mathcal{V}_{\text{M}}}\vert= \vert{\mathcal{V}}\vert \times K$ are the extended supra-nodes and $\mathcal{E}_{\text{M}} = \{(i,j) \in \mathcal{V}_{\text{M}} \times \mathcal{V}_{\text{M}}\}$ are the edges between supra-nodes. There are $K$ modality planes, each with adjacency matrices $\mathbf{A}_{(k)} \in \mathcal{R}^{P \times P}$. The $K\times K$ pairwise cross planar connections are given by $\mathbf{C}_{(l,m)} \in \mathcal{R}^{P \times P}$, where $P=\vert{\mathcal{V}}\vert$. All edge weights may take values in range $[0-1]$

\subsection{HGR Maximal Correlations for Latent Multi-Graph Learning:}
Recall that we would like to learn task informed patient-modality multi-graph representations automatically from unstructured modality data. To this end, we develop the framework illustrated in the Gray Box in Fig.~\ref{MaxCorrMGNN_Formulation}.

Let $\{\mathbf{x}_{n}^{k} \in \mathcal{R}^{D_{k} \times 1}\}^{K}_{k=1}$ be the features from to modality $k$ for patient $n$. Since features from different modalities lie in different input subspaces, we develop parallel common space projections to explore the dependence between them. The Hirschfield, Gebelin, R\`enyi (HGR)~\cite{wang2019efficient} framework in statistics is known to generalize the notion of dependence to abstract and non-linear functional spaces. Such non-linear projections can be parameterized by deep neural networks.

Specifically, let the collection of modality-specific projection networks be given by $\{\mathbf{f}^{k}(\cdot): \mathcal{R}^{D_{k}\times 1} \rightarrow \mathcal{R}^{D_{p}\times 1}\}$. The HGR maximal correlation is a symmetric measure obtained by solving the following coupled pairwise constrained optimization problem:

\begin{equation}
    \sup_{\mathcal{C}_{E},\mathcal{C}_{\text{Cov}}}  {\rho_{\text{HGR}}(\mathbf{x}^{l},\mathbf{x}^{m})} = \sup_{\mathcal{C}_{E},\mathcal{C}_{\text{Cov}}}  {\mathbb{E}\Big[[\mathbf{f}^{l}(\mathbf{x}^{l})]^{T}\mathbf{f}^{m}(\mathbf{x}^{m})\Big]} 
    \label{HGR}
\end{equation}
$\forall \{l.m\} \ \text{s.t.}  \ l \neq m$, where $\mathbf{\mathcal{I}}_{D_{p}}$ is a $D_p\times D_p$ identity matrix.
The constraint sets are given by:
\begin{eqnarray}
\mathcal{C}_{E} :  \{\mathbb{E}[\mathbf{f}^{l}(\cdot)]=
    \mathbb{E}[\mathbf{f}^{m}(\cdot)]=\mathbf{0} \} \ \ \ \ \label{Exp} \\ \mathcal{C}_{\text{Cov}} : \{\mathbf{Cov}(\mathbf{f}^{l}(\mathbf{x}^{l}))= \mathbf{Cov}(\mathbf{f}^{m}(\mathbf{x}^{m})) = \mathbf{\mathcal{I}}_{D_{p}}\} \label{Cov}
\end{eqnarray}
Approaches such as deep CCA~\cite{yang2019survey} can be thought of as a special case of this formulation which solve the whitening (empricial covariance) constraints (Eq.~(\ref{Cov})) via explicit pairwise de-correlation. 

However, for multiple modalities in large datasets, exact whitening is not scalable. To circumvent this issue, we can use the approach in~\cite{wang2019efficient}. This formulation proposes introduces a relaxation to the exact HGR, named soft-HGR, which consists of a trace regularizer in lieu of whitening. Eq.~(\ref{HGR}) can be relaxed as an empirical minimization problem $\text{min} \  \mathcal{L}_{\text{sHGR}}$, where the sHGR loss is:

\begin{gather}
    \mathcal{L}_{\text{sHGR}} = - \frac{1}{N_z} \mathbb{E}\Big[\mathbf{f}^{l}(\mathbf{x}^{l})^{T}\mathbf{f}^{m}(\mathbf{x}^{m})\Big] + \frac{1}{2N_z}\Tr\Big[{\mathbf{Cov}[\mathbf{f}^{l}(\mathbf{x}^{l})]}{\mathbf{Cov}[\mathbf{f}^{m}(\mathbf{x}^{m})]}\Big] \label{sHGR} \\
    = - \frac{1}{N_z} \Bigg( \sum^{K}_{l,m=1} \Bigg[ \sum^{N}_{n=1} \frac{\mathbf{f}^{l}(\mathbf{x}_{n}^{l})^{T}\mathbf{f}^{m}(\mathbf{x}_{n}^{m})}{(N-1)} - \frac{\Tr\Big[{\mathbf{Cov}[\mathbf{f}^{l}(\mathbf{x}^{l})]}{\mathbf{Cov}[\mathbf{f}^{m}(\mathbf{x}^{m})]}\Big]}{2} \Bigg]\Bigg) \ \ l \neq m \nonumber
\end{gather}
The expectation under the functional transformations  $\mathbb{E}[\mathbf{f}^{l}(\mathbf{x}^{l})]=\mathbb{E}[\mathbf{f}^{m}(\mathbf{x}^{m})]=\mathbf{0}$ is enforced step-wise by mean subtraction during optimization. Here, $\mathbf{Cov}(\cdot)$ is the empirical covariance matrix. We parameterize $\{\mathbf{f}^{k}(\cdot)\}$ as a simple two layered fully connected neural network with a normalization factor as $N_z = M(M-1)$. 

By design, the MaxCorr formulation allows us to utilize the correlation $\rho_{\text{sHGR}}(\mathbf{x}_{i}^{l},\mathbf{x}_{j}^m)$ (computed after solving Eq.~(\ref{sHGR})) to model dependence between patients $i$ and $j$ according to the $l$ and $m$ modality features in a general setting. The absolute value of this correlation measure define the edge weights between nodes in the patient-modality multi-graph. As opposed to existing

\paragraph{\textbf{Learnable Adaptive Sparsity:}} Additionally, we would like to have our learning framework automatically discover and retain salient edges that are relevant for prediction. To encourage sparsity in the edges, we utilize a learnable soft-thresholding formulation. We first define a symmetric block sparsity matrix $\mathbf{S}$. Since edge weights in the multi-graph are in the range $[0-1]$, we normalize it through the sigmoid function as $\tilde{\mathbf{S}}= \tilde{\mathbf{S}}^{T} = \text{Sigmoid}({\mathbf{S}}) \in \mathcal{R}^{K \times K}$. The entries of the soft-thresholding matrix $\tilde{\mathbf{S}}[l,m]$ define learnable thresholds for the cross modal connections when $l \neq m$ and in-plane connections when $l=m$. Finally, the cross modal edges and in-plane edges of the multi-graph are given by
\begin{eqnarray}
\mathbf{C}_{(l,m)}[i,j] = \text{ReLU}({\tilde{\rho}_{\text{sHGR}}(\mathbf{x}_{i}^{l},\mathbf{x}_{j}^{m})} - \tilde{\mathbf{S}}[l,m]) \\
\mathbf{A}_{k}[i,j] = \text{ReLU}(\tilde{\rho}_{\text{sHGR}}(\mathbf{x}_{i}^{k},\mathbf{x}_{j}^{k})- \tilde{\mathbf{S}}[k,k])
\end{eqnarray}
with ${\tilde{\rho}_{\text{sHGR}}} = \vert{\rho}_{\text{sHGR}}\vert$ respectively. The adjacency matrices $\mathbf{A}_{(k)}$ model the dependence within the features of modality $k$, while the cross planar matrices $\{\mathbf{C}_{(l,m)}\}$ capture interactions across modalities. Overall, $\mathbf{S}$ acts as a regularizer that suppresses noisy weak dependencies. These regularization parameters are automatically inferred during training along with the MaxCorr projection parameters $\{\mathbf{f}^{k}(\cdot)\}$. This effectively adds just $K(K+1)$ learnable parameters to the MaxCorrMGNN.

\subsection{Multi-Graph Neural Network:}

As a standalone optimization, the MaxCorr block is not guaranteed to learn discriminative projections of modality features. A natural next step is to couple the multi-graph representation learning with the classification task. Graph Neural Networks have recently become popular tools for reasoning from graphs and graph-signals. Given the patient-modality multi-graph, we design an extension of traditional graph neural networks to multi-graphs for inference tasks.

Conventional GNNs filter information based on the graph topology (i.e. the adjacency matrix) to map the node features to the targets based on graph traversals. Conceptually,
cascading $l$ GNN layers is analogous to filtered information pooling at each node from its $l$-hop neighbors~\cite{kipf2016semi} inferred from the powers of the graph adjacency matrix. These neighborhoods can be reached by seeding walks on the graph starting at the desired node. Inspired by this design, we craft a multi-graph neural network (Purple Box in Fig.~\ref{MaxCorrMGNN_Formulation}) for outcome prediction. Our MGNN generalizes structured message passing to the multi-graph in a manner similar to those done for multiplexed graphs~\cite{d2022fusing}. Notably, our formulation is more general, as it avoids using strictly vertical interaction constraints between patients across modalities.

We first construct two supra-adjacency matrices to perform walks on the multi-graph $\mathcal{G}_{\text{M}}$ for fusion. The first is the \textit{intra-modality adjacency matrix} $\boldsymbol{\mathcal{A}} \in \mathcal{R}^{PK \times PK}$. The second is the \textit{inter-modality connectivity matrix} ${\boldsymbol{\mathcal{C}}} \in \mathcal{R}^{PK \times PK}$, each defined block-wise. Mathematically, we express this as: 
\begin{gather}
\boldsymbol{\mathcal{A}}= \bigoplus_{k}\mathbf{A}_{(k)}\\
\hat{\boldsymbol{\mathcal{C}}} : \hat{\boldsymbol{\mathcal{C}}}[lP:(l+1)P,mP:(m+1)P] =  \mathbf{C}_{(l,m)} \ \mathbb{1}(l\neq m) 
+ \boldsymbol{\mathcal{I}_{P}} \ \mathbb{1}(l=m)  \label{ILtrans}
\end{gather}
where $\bigoplus$ is the direct sum operation and
$\mathbb{1}$ denotes the indicator function. By design, $\boldsymbol{\mathcal{A}}$ is block-diagonal and allows for within-planar (intra-modality) transitions between nodes. The off-diagonal blocks of $\hat{\boldsymbol{\mathcal{C}}}$, i.e. $\mathbf{C}_{(l,m)}$, capture transitions between nodes as per cross-planar (inter-modality) relationships.  

\paragraph{\textbf{MGNN Message Passing Walks:}} Walks on $ \mathcal{G}_{\text{M}}$ combine within and across planar steps to reach a patient supra-node $s_j$ from another supra-node $s_i$ ($s_i,s_j\in \mathcal{V}_{\text{M}}$). We characterize the multi-hop neighborhoods and transitions using factorized operations involving $\boldsymbol{\mathcal{A}}$ and $\hat{\boldsymbol{\mathcal{C}}}$. We can perform a multi-graph walk via two types of distinct steps, i.e., (1) an isolated intra-planar transition or (2) a transition involving an inter-planar step either before or after a within-planar step. These steps can be exhaustively recreated via two factorizations: (I) \textit{after} one intra-planar step, the walk \textit{may} continue in the same modal plane or hop to a different one via $\boldsymbol{\mathcal{A}}\hat{\boldsymbol{\mathcal{C}}}$ and (II) the walk \textit{may} continue in the current modal plane or hop to a different plane \textit{before} the intra-planar step via $\hat{\boldsymbol{\mathcal{C}}}\boldsymbol{\mathcal{A}}$. 

The Multi-Graph Neural Network (MGNN) uses these walk operations to automatically mine predictive patterns from the multi-graph given the targets (task-supervision) and the GNN parameters. For supra-node $s_i$, $\mathbf{h}^{(d)}_{s_i} \in \mathcal{R}^{D^{d}\times 1}$ is the feature (supra)-embedding at MGNN depth $d$. The forward pass operations of the MGNN are as follows:
\begin{eqnarray}
\mathbf{h}_{s_i,I}^{(d+1)} = \boldsymbol{\phi}^{(d)}_{I}\Big((1+\epsilon) \mathbf{h}^{(d)}_{s_i} + \text{wmean}\Big[\mathbf{h}^{(d)}_{s_j}, \boldsymbol{\mathcal{A}}\mathbf{\hat{\boldsymbol{\mathcal{C}}}}[s_i,s_j] \ \ ;\ \ s_j \in \mathcal{N}_{\boldsymbol{\mathcal{A}}\hat{\boldsymbol{\mathcal{C}}}}(s_i) \Big] \Big)   \label{I} \ \ \ \ \ \ \\
\mathbf{h}_{s_i,II}^{(d+1)} = \boldsymbol{\phi}^{(d)}_{II}\Big( (1+\epsilon)\mathbf{h}^{(d)}_{s_i} + \text{wmean}\Big[\mathbf{h}^{(d)}_{s_j}, \mathbf{\hat{\boldsymbol{\mathcal{C}}}\boldsymbol{\mathcal{A}}}[s_i,s_j] \ \ ; \ \ s_j \in \mathcal{N}_{\hat{\boldsymbol{\mathcal{C}}}\boldsymbol{\mathcal{A}}}(s_i) \Big] \Big) \ \ \ \ \ \  \label{II} \\ 
\mathbf{h}^{(d+1)}_{s_i} = \text{concat}(\mathbf{h}^{(d+1)}_{s_i,I},\mathbf{h}^{(d+1)}_{s_i,II}) \ \ \ \ \  \ \ \ \ \ \ \ \ \ \ \ \ \ \ \ \ \ \ \\  \ \ \ \ \ \ \ \ \ \ \ \ \ \ \mathbf{g}_{o}(\{\mathbf{h}^{(L)}_{s_i}\}_{s_i \leftrightarrow i}) = \hat{\mathbf{Y}}_{i} \ \ \ \ \ \ \ \ \ \ \ \ \ \ \ \ \ \ \ \ \ \ \ \ \ \ \ \label{MultiMP}
\end{eqnarray}
At the input layer, we have $\mathbf{h}_{s_i}^{(0)} = \mathbf{f}^{k}(\mathbf{x}_{i}^{k})$ computed from the modality features for patient $i$ after the sHGR transformation from the corresponding modality $k$. We then concatenate the supra-embeddings as input to the next layer i.e. $\mathbf{h}^{(d+1)}_{s_i}$. Eqs.~(\ref{I}-\ref{II}) denote the Graph Isomorphism Network (GIN)~\cite{xu2018powerful} with $\{\boldsymbol{\mathcal{\phi}}^{(d)}_{I}(\cdot) , \boldsymbol{\mathcal{\phi}}^{(d)}_{II}(\cdot)\}$ as layerwise linear transformations. This performs message passing on the multi-graph using the neighborhood relationships and normalized edge weights from the walk matrices in the weighted mean operation  $\text{wmean}(\cdot)$. 

From the interpretability standpoint, these \textit{operations keep the semantics of the embeddings intact at both the patient and modality level throughout the MaxCorrMGNN transformations}. Finally, $\mathbf{g}_{o}(\cdot)$ is a graph readout network that maps to the one-hot encoded outcome $\mathbf{Y}$, which performs a convex combination of the filtered modality embeddings, followed by a linear readout. 

\subsection{End-to-end Learning through task supervision:}
Piecing together the constituent components, i.e. the latent graph learning and MGNN inference module, we optimize the following coupled objective function:
\begin{equation}
\mathcal{L} = \lambda \mathcal{L}_{\text{sHGR}} + (1-\lambda) \mathcal{L}_{\text{CE}}(\hat{\mathbf{Y}},\mathbf{Y})
\label{loss}
\end{equation}
with $\lambda \in [0,1]$ being a tradeoff parameter and $\mathcal{L}_{CE}(\cdot)$ being the cross entropy loss. The parameters $\{\{\mathbf{f}^{k}(\cdot)\},\mathbf{S},\{\boldsymbol{\mathcal{\phi}}^{(d)}_{I}(\cdot), \boldsymbol{\mathcal{\phi}}^{(d)}_{II}(\cdot),\epsilon\},\mathbf{g}_{o}(\cdot)\}$ of the framework are jointly learned via standard backpropagation. 

\paragraph{{\textbf{Inductive Learning for Multi-Graph Neural Networks:}}}
The multi-graph is designed to have subjects as the nodes, which requires us to adapt training to accomodate an inductive learning setup. Specifically, we train the MaxCorrMGNN in a fully supervised fashion by extending the principles outlined in ~\cite{cosmo2020latent} for multi-layered graphs. During training, we use only the supra-node features and induced sub-graph edges (including both cross-modal and intra-planar edges) associated with the subjects in the training set for backpropagation. During validation/testing, we freeze the parameter estimates and add in the edges corresponding to the unseen patients to perform a forward pass for estimation. This procedure ensures that no double dipping occurs in the hyper-parameter estimation, nor in the evaluation step. Additionally, while not the focus of this application, this procedure allows for extending prediction and training to an online setting, where new subject/modality information may dynamically become available.

\paragraph{{\textbf{Implementation Details:}}}
We implement the MaxCorr projection networks $\mathbf{f}^{l}(\cdot)$ as a simple three layered neural network with hidden layer width of $32$ and output $D_{p} = 64$ and LeakyReLU activation (negative slope=0.01). The MGNN layers are Graph Isomorphism Network (GIN)\cite{xu2018powerful} with ReLU activation and linear readout (width:$64$) and batch normalization. $\mathbf{g}_{o}(\cdot)$ implements a convex combination of the modality embeddings followed by a linear layer. We use the ADAMw optimizer~\cite{loshchilov2017decoupled} and train on a 64GB CPU RAM, 2.3 GHz 16-Core Intel i9 machine (18-20 min training time per run). We set the hyperparameters for our model (and baselines) using grid-search to $\lambda=0.01$, learning rate$=0.0001$, weight decay$=0.001$, epochs$=50$, batch size$=128$ after pre-training the network on the sHGR loss alone for $50$ epochs. All frameworks are implemented on the Deep Graph Library (v=0.6.2) in PyTorch (v=0.10.1). 

\section{Experiments and Results}
\label{Data}

\subsection{{Data and Preprocessing}} 
\begin{figure}[t!]
\begin{center}
\centerline
{\includegraphics[scale=0.33]{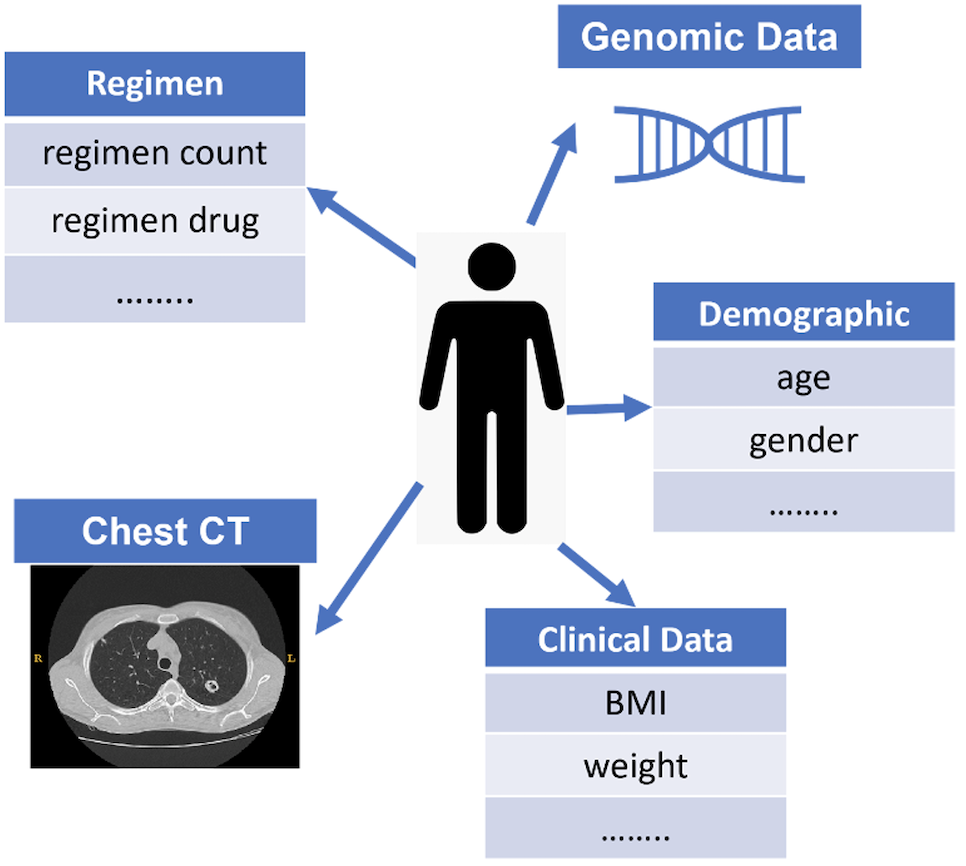}}
\caption{{NIH TB Dataset: Multimodal data for treatment outcome prediction.}}
\label{TB}
\end{center}
\end{figure}

We evaluate our model on the Tuberculosis Data Exploration Portal \cite{Gabrielian2019} consisting of 3051 patients with five different treatment outcomes (Died, Still on treatment, Completed, Cured, or Failure) with the class frequencies as: 0.21/0.11/0.50/0.10/0.08 respectively and five modalities. We pre-process the data according to the procedure outlined in~\cite{d2022fusing}.

For each subject, we have features available from demographic, clinical, regimen and genomic recordings with chest CTs available for 1015 of them. We have a total of 4081 genomic, 29 demographic, 1726 clinical, 233 regimen features that are categorical, and 2048 imaging and 8 miscellaneous continuous features. Information that may directly be related to treatment outcomes, eg drug resistance type, were removed from the clinical and regimen features.

For genomic data, 81 single nucleotide polymorphisms (SNPs) from the causative organisms \textit{Mycobacterium tuberculosis} (Mtb) known to be related to drug resistance were used. For 275 of the subjects, we also assemble the raw genome sequence from NCBI Sequence Read Archive. This provides a more fine-grained description of the biological sequences of the causative pathogen~\cite{Seabolt2019}. Briefly, we performed a \textit{de novo} assembly process on each Mtb genome to yield protein and gene sequences. We utilized InterProScan~\cite{Jones2014} to further process the protein sequences and extract the functional domains, i.e. sub-sequences located within the protein's amino acid chain responsible for the enzymatic bioactivity of a protein. This provides a total of 4000 functional genomic features.  Finally, for the imaging modality, the lung was segmented via multi-atlas segmentation~\cite{wang2013multi} followed by a pre-trained DenseNet \cite{huang2017densely} to extract a 1024-dimensional feature vector for each axial slice intersecting the lung. The mean and maximum of each feature were then assembled to give a total of 2048 features. Missing features are imputed from the training cohort using mean imputation for all runs.

\subsection{{Evaluation Metrics:}} Since we have a five-class classification task, we evaluate the prediction performance of the MaxCorrMGNN and the baselines using the AU-ROC (Area Under the Receiver Operating Curve) metric. Given the prediction logits, this metric is computed both class-wise and as a weighted average. Higher per-class and overall AU-ROC indicate improved performance. For our experiments, we use $10$ randomly generated train/validation/test splits with ratio 0.7/0.1/0.2 to train our model and each baseline. 

Finally, statistical differences between the baselines and our method are measured according to the DeLong \cite{delong1988comparing} test computed class-wise. This test is a sanity check to evaluate whether perceived differences in model performance are robust to sampling.

\subsection{Baseline Comparisons}

We perform a comprehensive evaluation of our framework for the problem of multimodal fusion. Our baseline comparisons can be grouped into three categories, namely, (1) Single Modality Predictors/ No Fusion (2) State-of-the-art Conventional including early/late/intermediate fusion and Latent-Graph Learning models from literature (3) Ablation Studies. 

The ablation studies evaluate the efficacy of the three main constituents of the MaxCorrMGNN, i.e. the MaxCorr graph construction, the Multi-Graph Neural Network and the end-to-end optimization.

\begin{itemize}

\item  \textbf{Single Modality:} For this comparison, we run predictive deep-learning models on the individual modality features without fusing them as a benchmark. We use a two layered multi-layered perception (MLP)  with hidden layer widths as 400 and 20 and LeakyReLU activation (neg. slope=0.01).

\medskip
\item \textbf{Early Fusion:} For early fusion, individual modality features are first concatenated and then fed through a neural network. The predictive model has the same architecture as the previous baseline. 

\medskip
\item \textbf{Uncertainty Based Late Fusion~\cite{wang2021modeling}:} We combine the predictions from the individual modalities in the previous baseline using a state-of-the-art late fusion framework in~\cite{wang2021modeling}. This model estimates the uncertainty in the individual classifiers to improve the robustness of outcome prediction. Unlike our work, patient-modality dependence is not explicitly modeled as the modality predictions are only combined after individual modality-specific models have been trained. Hyperparameters are set according to \cite{wang2021modeling}. 

\medskip
\item \textbf{Graph Based Intermediate Fusion~\cite{d2022fusing}:} This is a graph based neural framework that achieved state-of-the-art performance on multimodal fusion on unstructured data. This model follows a two step procedure. For each patient, this model first converts the multimodal features into a fused binary multiplex graph (multi-graph where all blocks of $\mathbf{\hat{\boldsymbol{\mathcal{C}}}}$ are strictly diagonal) between features. The graph connectivity is learned in an unsupervised fashion through auto-encoders. Following this, a multiplexed graph neural network is used for inference.  Hyperparameters are set according to \cite{d2022fusing}. While this framework takes a graph based approach to fusion, the construction of the graph is not directly coupled with the task supervision.

\medskip
\item \textbf{Latent Graph Learning~\cite{cosmo2020latent}:} This baseline was developed for fusing multimodal data for prediction. It introduces a latent patient-patient graph learning from the concatenated modality features via a graph-attention (GAT-like~\cite{velivckovic2017graph}) formulation. However, unlike our model, the feature concatenation does not distinguish between intra- and inter-modality dependence across patients i.e. it constructs a single-relational (collapsed) graph that is learned as a part of the training. 

\medskip
\item \textbf{sHGR+ANN~\cite{wang2019efficient}:} This is a state-of-the-art multimodal fusion framework~\cite{wang2019efficient} that also utilizes the sHGR formulation to infer multi-modal data representations. However, instead of constructing a patient-modality graph, the projected features are combined via concatenation. Then, a two layered MLP (hidden size:200) maps to the outcomes, with the two objectives trained end-to-end. This baseline can be thought of as an \textit{ablation} that evaluates the benefit of using the multi-graph neural network for fine-grained reasoning. Additionally, this and the previous framework help us evaluate the benefit of our patient-modality multi-graph representation for fusion. 

\medskip
\item \textbf{MaxCorrMGNN w/o sHGR:} Through this comparison, we evaluate the need for using the soft HGR formulation to construct the latent multi-graph. Keeping the architectural components consistent with our model, we set $\lambda=0$ in Eq.~(\ref{loss}). Note that this \textit{ablation} effectively converts the multi-graph representation learning into a modality specific self/cross attention learning, akin to graph transformers. Overall, this framework helps us evaluate the benefit of our MaxCorr formulation for latent multi-graph learning. 

\medskip
\item \textbf{Decoupled MaxCorrMGNN:} Finally, this \textit{ablation} is designed to examine the benefit of coupling the MaxCorr and MGNN into a coupled objective. Therefore, instead of an end-to-end training, we run the sHGR optimization first, followed by the MGNN for prediction.
\end{itemize}

\begin{figure*}[t!]
\begin{center}
\centerline{\includegraphics[width=\textwidth]{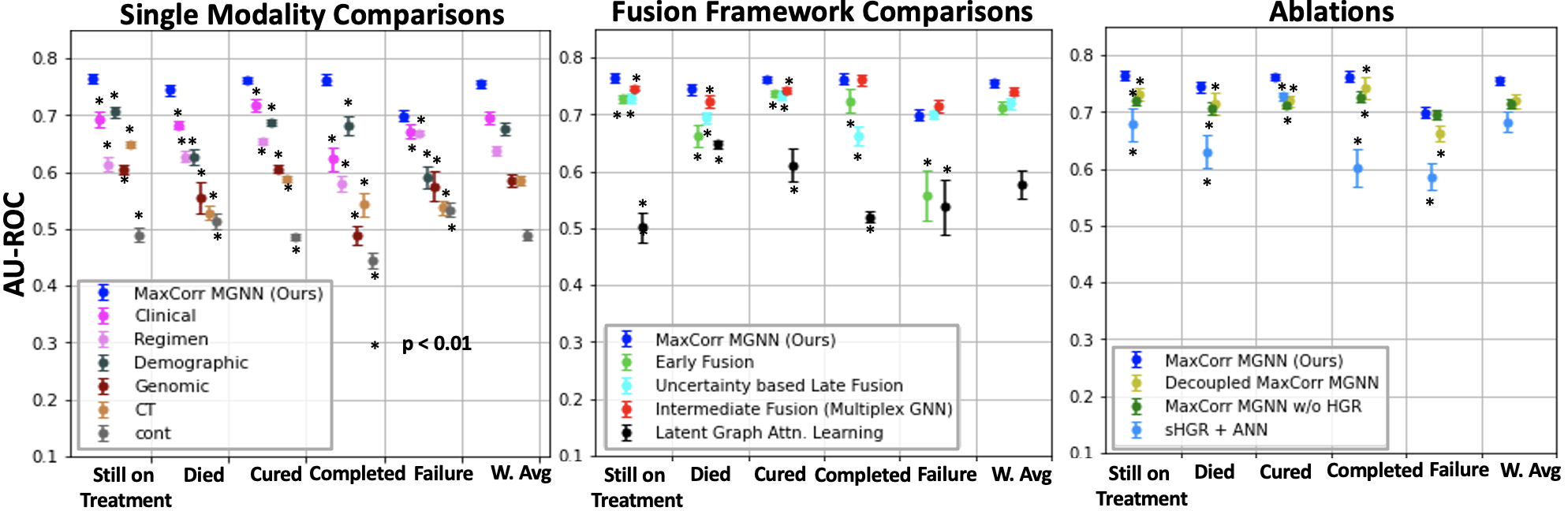}}
\caption{We display the mean per-class and weighted average AU-ROC and the standard errors for TB outcome prediction against \textbf{(Left)}: Single Modality Predictors \textbf{(Middle)}: Traditional and Graph Based Fusion Frameworks \textbf{(Right)}: Ablations of the MaxCorrMGNN. * indicate comparisons against the MaxCorrMGNN  according to the DeLong test that achieve statistical significance ($p<0.01$).}
\label{TB Results}
\end{center}
\end{figure*}
\subsection{{Outcome Prediction Performance:}} 

Fig.~\ref{TB Results} illustrates the outcome prediction performance of our framework against the single modality predictors (left), state-of-the-art fusion frameworks (middle), and  ablated versions of our model (right). Comparisons marked with $*$ achieve a statistical significance threshold of $p<0.01$ across runs as per the DeLong test~\cite{delong1988comparing}. Note that our fusion framework outperforms all of the single modality predictors by a large margin. Moreover, the traditional and graph-based fusion baselines also provide improved performance against the single modality predictors. Taken together, these observations highlight the need for fusion of multiple modalities for outcome prediction in TB. This observation is consistent with findings in treatment outcome prediction literature~\cite{d2022fusing,munoz2010factors} in TB.

The MaxCorrMGNN also provides improved performance when compared to all of the fusion baselines, with most comparisons achieving statistical significance thresholds. While the Early Fusion and Uncertainty based Late fusion~\cite{wang2021modeling} networks provide marked improvements over single modality predictions, but still fail to reach the performance level of our model. This is likely due to their limited ability to leverage subtle patient-specific cross-modal interactions. 

On the other hand, the latent graph learning in~\cite{cosmo2020latent} models connectivity between subjects as a part of the supervision. However, this method collapses the different types of dependence into one relation-type, which may be too restrictive for fusion applications. The intermediate fusion framework of~\cite{d2022fusing} was designed to address these limitations by the use of multiplex graphs. However, the artificial separation between the graph construction and inference steps may not inherently extract discriminative multi-graph representations, which could explain the performance gap against our framework. 

Finally, the three ablations, the sHGR+ANN~\cite{wang2019efficient}, MaxCorrMGNN w/o sHGR, Decoupled MaxCorrMGNN help us systematically examine the three building blocks of our framework, i.e. MGNN and MaxCorr networks individually as well as the end-to-end training of the two blocks. We observe a notable performance drop in these baselines, which reinforces the principles we considered in carefully designing the individual components. In fact, the comparison against the Decoupled MaxCorrMGNN illustrates that coupling the two components into a single objective is key to obtaining improved representational power for predictive tasks. Taken together, our results suggest that the MaxCorrMGNN is a powerful framework for multimodal fusion of unstructured data.

\section{Discussion}
We have developed a novel multi-graph deep learning framework, i.e. the MaxCorrMGNN for generalized problems of multimodal fusion in medical data. Going one step beyond simple statistical measures, the patient-modality multi-layered graph allows us to uncover nuanced non-linear notions of dependence between modality features via the maximal correlation soft-HGR formulation. The sHGR formulation coupled with the learnable sparsity module allow us to directly translate an abstract measure of interaction across subjects and modalities in any multimodal dataset into a patient-modality multi-layered graph structure for inference. The construction of the multi-graph planes allow the node features to retain their individuality in terms of the plane (modality) and patient (node-identity) in the filtered Graph Neural Network representations. This admits more explainable intermediate representations in comparison to the baselines, i.e. provides us with the ability to explicitly reason at the granularity of both the subjects and modalities. Conversely, the graph based/traditional fusion baselines collapse this information, either in the multimodal representation or in the inference step.  We believe that this added flexibility in the MaxCorrMGNN contributes to the improved generalization power in practice. Finally, all the individual components (i.e. MaxCorr, learnable soft-thresholding, MGNN message passing) are designed to be fully differentiable deep learning operations, allowing us to directly couple them end-to-end. We demonstrate in experiment that this coupling is key to generalization. As such, this model makes very mild assumptions about the nature of the multimodal data. The general principles and machinery developed in this work would likely be useful to a wide variety of applications beyond the medical realm.

\paragraph{\textbf{Limitations and Future Work:}}
In problems of multimodal fusion, especially for medical applications, data acquisition is a fairly contrived and expensive process. In many real-world modalities may often be only partially observed, missing in totality, or noisy in acquisition. Simple methods such as mean based imputation may be inadequate for fine-grained reasoning. As an aim to address this, an active line of exploration is to extend the framework to handle missing, ambiguous and erroneous data and labels within the multilayered graph representation. This may be achieved by leveraging statistical and graph theoretic tools that can be integrated directly into the message passing walks. Finally, the multi-graph and HGR construction focuses on uncovering pairwise relationships between subjects and features. A future direction would be to extend these frameworks to model complex multi-set dependencies.

\section{Conclusion}
We have introduced a novel multi-layered graph based neural framework for general inference problems in multimodal fusion. Our framework leverages the HGR MaxCorr formulation to convert unstructured multi-modal data into a patient-modality multi-graph. We design a generalized multi-graph neural network for fine-grained reasoning from this representation. Our design preserves the patient-modality semantics as a part of the architecture, making our representations more readily interpretable rather than fully black-box. The end-to-end optimization of the two components offers a viable tradeoff between flexibility, representational power, and interpretability. We demonstrate the efficacy of the MaxCorr MGNN for fusing disparate information from imaging, genomic and clinical data for outcome prediction in Tuberculosis and demonstrate consistent improvements against competing state-of-the-art baselines developed in literature.  Moreover, the framework makes very few assumptions making it potentially applicable to a variety of fusion problems in other AI domains. Finally, the principles developed in this paper are general and can potentially be applied to problems well beyond multimodal fusion.
\bibliographystyle{splncs04}
\bibliography{paper_0033.bib}

\begin{thebibliography}{10}
\providecommand{\url}[1]{\texttt{#1}}
\providecommand{\urlprefix}{URL }
\providecommand{\doi}[1]{https://doi.org/#1}

\bibitem{baltruvsaitis2018multimodal}
Baltru{\v{s}}aitis, T., Ahuja, C., Morency, L.P.: Multimodal machine learning:
  A survey and taxonomy. IEEE transactions on pattern analysis and machine
  intelligence  \textbf{41}(2),  423--443 (2018)

\bibitem{cosmo2020latent}
Cosmo, L., Kazi, A., Ahmadi, S.A., Navab, N., Bronstein, M.: Latent-graph
  learning for disease prediction. In: Medical Image Computing and Computer
  Assisted Intervention--MICCAI 2020: 23rd International Conference, Lima,
  Peru, October 4--8, 2020, Proceedings, Part II 23. pp. 643--653. Springer
  (2020)

\bibitem{Cozzo2018}
Cozzo, E., de~Arruda, G.F., Rodrigues, F.A., Moreno, Y.: Multiplex networks
  (2018). \doi{10.1007/978-3-319-92255-3},
  \url{http://link.springer.com/10.1007/978-3-319-92255-3}

\bibitem{delong1988comparing}
DeLong, E.R., DeLong, D.M., Clarke-Pearson, D.L.: Comparing the areas under two
  or more correlated receiver operating characteristic curves: a nonparametric
  approach. Biometrics pp. 837--845 (1988)

\bibitem{dsouza2021m}
Dsouza, N.S., Nebel, M.B., Crocetti, D., Robinson, J., Mostofsky, S.,
  Venkataraman, A.: M-gcn: A multimodal graph convolutional network to
  integrate functional and structural connectomics data to predict
  multidimensional phenotypic characterizations. In: Medical Imaging with Deep
  Learning. pp. 119--130. PMLR (2021)

\bibitem{d2022fusing}
D’Souza, N.S., Wang, H., Giovannini, A., Foncubierta-Rodriguez, A., Beck,
  K.L., Boyko, O., Syeda-Mahmood, T.: Fusing modalities by multiplexed graph
  neural networks for outcome prediction in tuberculosis. In: Medical Image
  Computing and Computer Assisted Intervention--MICCAI 2022: 25th International
  Conference, Singapore, September 18--22, 2022, Proceedings, Part VII. pp.
  287--297. Springer (2022)

\bibitem{d2021matrix}
D’Souza, N.S., Nebel, M.B., Crocetti, D., Robinson, J., Mostofsky, S.,
  Venkataraman, A.: A matrix autoencoder framework to align the functional and
  structural connectivity manifolds as guided by behavioral phenotypes. In:
  Medical Image Computing and Computer Assisted Intervention--MICCAI 2021: 24th
  International Conference, Strasbourg, France, September 27--October 1, 2021,
  Proceedings, Part VII 24. pp. 625--636. Springer (2021)

\bibitem{d2021deep}
D’Souza, N.S., Nebel, M.B., Crocetti, D., Robinson, J., Wymbs, N., Mostofsky,
  S.H., Venkataraman, A.: Deep sr-ddl: Deep structurally regularized dynamic
  dictionary learning to integrate multimodal and dynamic functional
  connectomics data for multidimensional clinical characterizations. NeuroImage
   \textbf{241},  118388 (2021)

\bibitem{d2020deep}
D’Souza, N.S., Nebel, M.B., Crocetti, D., Wymbs, N., Robinson, J., Mostofsky,
  S., Venkataraman, A.: A deep-generative hybrid model to integrate multimodal
  and dynamic connectivity for predicting spectrum-level deficits in autism.
  In: Medical Image Computing and Computer Assisted Intervention--MICCAI 2020:
  23rd International Conference, Lima, Peru, October 4--8, 2020, Proceedings,
  Part VII 23. pp. 437--447. Springer (2020)

\bibitem{Gabrielian2019}
Gabrielian, A., Engle, E., Harris, M., Wollenberg, K., Juarez-Espinosa, O.,
  Glogowski, A., Long, A., Patti, L., Hurt, D.E., Rosenthal, A., Tartakovsky,
  M.: Tb depot (data exploration portal): A multi-domain tuberculosis data
  analysis resource. PLOS ONE  \textbf{14}(5),  e0217410 (may 2019).
  \doi{10.1371/journal.pone.0217410},
  \url{http://dx.plos.org/10.1371/journal.pone.0217410}

\bibitem{huang2017densely}
Huang, G., Liu, Z., Van Der~Maaten, L., Weinberger, K.Q.: Densely connected
  convolutional networks. In: Proceedings of the IEEE conference on computer
  vision and pattern recognition. pp. 4700--4708 (2017)

\bibitem{Jones2014}
Jones, P., Binns, D., Chang, H.Y., Fraser, M., Li, W., McAnulla, C., McWilliam,
  H., Maslen, J., Mitchell, A., Nuka, G., Pesseat, S., Quinn, A.F.,
  Sangrador-Vegas, A., Scheremetjew, M., Yong, S.Y., Lopez, R., Hunter, S.:
  {InterProScan 5: genome-scale protein function classification.}
  Bioinformatics (Oxford, England)  \textbf{30}(9),  1236--40 (may 2014).
  \doi{10.1093/bioinformatics/btu031}

\bibitem{kipf2016semi}
Kipf, T.N., Welling, M.: Semi-supervised classification with graph
  convolutional networks. arXiv preprint arXiv:1609.02907  (2016)

\bibitem{lahat2015multimodal}
Lahat, D., Adali, T., Jutten, C.: Multimodal data fusion: an overview of
  methods, challenges, and prospects. Proceedings of the IEEE  \textbf{103}(9),
   1449--1477 (2015)

\bibitem{loshchilov2017decoupled}
Loshchilov, I., Hutter, F.: Decoupled weight decay regularization. arXiv
  preprint arXiv:1711.05101  (2017)

\bibitem{munoz2010factors}
Mu{\~n}oz-Sellart, M., Cuevas, L., Tumato, M., Merid, Y., Yassin, M.: Factors
  associated with poor tuberculosis treatment outcome in the southern region of
  ethiopia. The International Journal of Tuberculosis and Lung Disease
  \textbf{14}(8),  973--979 (2010)

\bibitem{Seabolt2019}
Seabolt, E.E., Nayar, G., Krishnareddy, H., Agarwal, A., Beck, K.L.,
  Terrizzano, I., Kandogan, E., Roth, M., Mukherjee, V., Kaufman, J.H.:
  {OMXWare, A Cloud-Based Platform for Studying Microbial Life at Scale}  (nov
  2019), \url{http://arxiv.org/abs/1911.02095}

\bibitem{subramanian2020multimodal}
Subramanian, V., Do, M.N., Syeda-Mahmood, T.: Multimodal fusion of imaging and
  genomics for lung cancer recurrence prediction. In: 2020 IEEE 17th
  International Symposium on Biomedical Imaging (ISBI). pp. 804--808. IEEE
  (2020)

\bibitem{subramanian2021multi}
Subramanian, V., Syeda-Mahmood, T., Do, M.N.: Multi-modality fusion using
  canonical correlation analysis methods: Application in breast cancer survival
  prediction from histology and genomics. arXiv preprint arXiv:2111.13987
  (2021)

\bibitem{velivckovic2017graph}
Veli{\v{c}}kovi{\'c}, P., Cucurull, G., Casanova, A., Romero, A., Lio, P.,
  Bengio, Y.: Graph attention networks. arXiv preprint arXiv:1710.10903  (2017)

\bibitem{wang2021modeling}
Wang, H., Subramanian, V., Syeda-Mahmood, T.: Modeling uncertainty in
  multi-modal fusion for lung cancer survival analysis. In: 2021 IEEE 18th
  International Symposium on Biomedical Imaging (ISBI). pp. 1169--1172. IEEE
  (2021)

\bibitem{wang2013multi}
Wang, H., Yushkevich, P.: Multi-atlas segmentation with joint label fusion and
  corrective learning—an open source implementation. Frontiers in
  neuroinformatics  \textbf{7}, ~27 (2013)

\bibitem{wang2019efficient}
Wang, L., Wu, J., Huang, S.L., Zheng, L., Xu, X., Zhang, L., Huang, J.: An
  efficient approach to informative feature extraction from multimodal data.
  In: Proceedings of the AAAI Conference on Artificial Intelligence. vol.~33,
  pp. 5281--5288 (2019)

\bibitem{xu2018powerful}
Xu, K., Hu, W., Leskovec, J., Jegelka, S.: How powerful are graph neural
  networks? arXiv preprint arXiv:1810.00826  (2018)

\bibitem{yang2019survey}
Yang, X., Liu, W., Liu, W., Tao, D.: A survey on canonical correlation
  analysis. IEEE Transactions on Knowledge and Data Engineering
  \textbf{33}(6),  2349--2368 (2019)

\bibitem{zheng2022multi}
Zheng, S., Zhu, Z., Liu, Z., Guo, Z., Liu, Y., Yang, Y., Zhao, Y.: Multi-modal
  graph learning for disease prediction. IEEE Transactions on Medical Imaging
  \textbf{41}(9),  2207--2216 (2022)

\end{thebibliography}
\end{document}